# Spoken Language Translation for Polish

Krzysztof Marasek, Krzysztof Wołk, Danijel Koroinek, Łukasz Brocki, Ryszard Gubrynowicz Multimedia Department, Polish-Japanese Institute of Information Technology, 02-008 Warszawa, Koszykowa 86, Poland.

**Abstract.** Spoken language translation (SLT) is becoming more important in the increasingly globalized world, both from a social and economic point of view. It is one of the major challenges for automatic speech recognition (ASR) and machine translation (MT), driving intense research activities in these areas. While past research in SLT, due to technology limitations, dealt mostly with speech recorded un- der controlled conditions, today's major challenge is the translation of spoken language as it occurs in real life. Considered application scenarios range from portable translators for tourists, lectures and presentations translation, to broadcast news and shows with live captioning. We would like to present PJIIT's experiences in the SLT gained from the Eu-Bridge 7th framework project and the U-Star consortium activities for the Polish/English language pair. Presented research concentrates on ASR adaptation for Polish (state-of-the-art acoustic models: DBN-BLSTM training, Kaldi: LDA+MLLT+SAT+MMI), language modeling for ASR and MT (text normalization, RNN-based LMs, n-gram model domain interpolation) and statistical translation techniques (hierarchical mod-els, factored translation models, automatic casing and punctuation, comparable and bilingual corpora preparation). While results for the well-de ned domains (phrases for travellers, parliament speeches, medical documentation, movie subtitling) are very encouraging, less de ned domains (presentation, lectures) still form a challenge. Our progress in the IWSLT TED task (MT only) will be presented, as well as current progress in the Polish ASR.

## 1. Introduction

Spoken language translation (SLT) is becoming more and more important in the increasingly globalized world, both from a social and economic point of view. It is one of the major challenges for automatic speech recognition (ASR) and machine translation (MT), driving intense research activities in these areas. While past research in SLT, due to technology limitations, dealt mostly with speech recorded under controlled conditions, today's major challenge is the translation of spoken language as it can be found in real life. Considered application scenarios range from portable translators for tourists, lectures and presentations translation, to broadcast news and shows with live captioning.

Polish, one of the West-Slavic languages, due to its complex infection and free word order forms a challenge for statistical machine translation (SMT) as well as for automatic speech recognition (ASR). Seven cases, three genders, animate and inanimate nouns, adjectives agreed with nouns in terms of gender, case and number and a lot of words borrowed from other languages which are sometimes infected as the of Polish origin, cause problems in establishing vocabularies of manageable size for ASR and translation to/from other languages and sparseness of data for statistical model training.

The state of speech and language technologies in Polish is still weak compared to other languages [1], even though it is improving at a fast pace and should not be considered as an under-resourced language for very long. Several research projects have emerged in the last couple of years dealing with the topics of automation in the telephony environment [2], transcription of legal documents [3] and recently speech-to- speech translation in different settings [4].

Commercially, there have been a few local startups and a few attempts by world market leaders, but none have yet achieved real adaptation of LVCSR in the field, with the exception of a few (Google and Apple/Nuance) which include ASR as a free service with their other products.

Despite the ca. 60 millions of Polish speakers world- wide the number of publicly available resources for the preparation of SLT systems is rather limited, making the progress in that slower than for other languages. In this paper our e orts in preparation of the Polish to English SLT system for various tasks ranging from tourist to captioning and lectures are presented.

The remainder of the paper is structured as follows. In section 2 Polish data preparation is described; section 3 deals with parallel data, in sections 4 and 5 training of the ASR and SMT is presented. In 6 a description of the live speech-to-speech system is given. Finally, the paper concludes on a discussion on en- countered issues and future work perspectives in section 7.

## 2. Polish data preparation

### 2.1. Acoustic data

Acoustic corpora are a set of audio recording with corresponding transcription describing its content as accurately as possible. The transcription need not be time aligned with the audio, although it helps if it is aligned, at least to the sentence level. Another big advantage is speaker annotation, which helps with the adaptation process.

Such data is usually not easily available for most languages. Since it is quite costly to transcribe audio, very little of it is available for free. A couple of paid Polish corpora exist in language repositories, like LDA or ELRA and recently a few have been released for free under the MetaShare project. Unfortunately, audio data is very domain dependent - variance in acoustic environment (reverberation and noise conditions) and speech styles (number of speakers and their manner of speech) greatly affect the ASR performance. That is why in-domain data is always encouraged, at least as a boosting mechanism, if little data is available. The quantity of the data does play an important role. A good corpus size is measured in at least tens or hundreds of hours, but significant improvements were noticed with as much as 2000 hours [5].

We used Speecon [6] as a general corpus for the initial acoustic models. It is a large nearly-studio quality corpus with people from various regions in Poland reading short sentences and phrases. The data included in the training set consisted of about 40 hours of speech from 133 different speakers. This was coupled with a similar corpus recorded in the studio at our institute with approximately 28 hours and 221 speakers.

The projects we participated in provided some clear specifications of domains that are to be researched. The parliamentary data was acquired from the publicly available data on the government websites. This data was proofread and corrected by human transcribers, because the publicly available transcripts were fixed for presentation purposes and did not accurately describe the recorded speech. The size of this corpus was over 100 hours with almost 900 different speakers (not balanced).

The subtitling domain was trained using data acquired from the Euronews channel. These were short (1-5 minute long) reports on various news related topics that amounted to around 60 hours of automatically aligned (not proofread) data. Additionally, a smaller set of interview recordings were downloaded from a popular radio station website. This data was very useful because it contained fairly well speaker annotated transcripts of the interviews. The size of the down- loaded data amounted to approximately 28 hours and 159 speakers.

The lecture domain was the most di cult to obtain and most di cult to recognize. A dataset was created from lectures lectures recorded during the Nomadic project[1] at our institute, which produced about 13 hours of IT related lectures from 6 lecturers. More data was being prepared from other sources online as this paper was written.

## 2.2. Monolingual text data

To train Language Models (LMs), a large collection of textual data is needed. LMs are required for both ASR and SMT. In ASR, they are used to evaluate hypotheses generated by the decoder, while in SMT they are used for both languages in the language pair: to analyse and synthesize grammatically correct sentences.

The size of the language corpus should be as big as possible. Sizes of hundreds of millions and even billions of tokens are not uncommon for these tasks. The main goal is to represent as many words in as many contexts in order to achieve substantial statistical significance. The size is also highly correlated with the target vocabulary. Given the infected nature of the Polish language, the vocabulary for the equivalent problem in, for example English, is several times larger. Unfortunately, the amount of digitized textual data in Polish is also far smaller than in English. This makes the quality of such systems even worse.

All the language models were trained both well known sources like the IPI PAN corpus [7]

---

[1] http://nomad.p jwstk.edu.pl/

and various other online sites and publications: major newspapers, e-books, subtitles, etc. We also used the transcripts of the audio corpora for ne-tuning of the data for specific domains. Ultimately, however, more data al- ways proved to be beneficial in almost every case. The largest dataset used to train a LM contained over 145 million tokens in a raw text le that was ca. 1 GB in size.

## 3. Parallel data preparation

A parallel corpus contains the same textual data in two or more languages. There are several issues with such corpora, and just like in the case of acoustic corpora, alignment of parallel streams of data plays a significant role in the quality of training. Parallel corpora are generally difficult to obtain and very expensive to produce - even more than acoustic corpora.

In our experiments three main types of corpora were used. First two belonged to very narrow domains: the European parliament Proceedings and medical texts. Second was a system based on movie subtitles, because they can be considered as a sample of natural human dialogs. Finally, a system with a wider domain, based on TED lectures on many different topics was also prepared.

The Polish data in the TED talks (15 MB) included almost 2 million words that were not tokenized [8]. The transcripts themselves were provided as pure text encoded in UTF-8 as prepared by the FBK team. Additionally, they were separated into sentences (one per line) and aligned in language pairs. A substantial amount (90 MB) of the English data included the PL-EN Europarl v7 corpus prepared during the Euromatrix project. We also used the 500 MB OPUS OpenSubtutles'12 corpus created from a collection of documents obtained from movie subtitles. The EMEA corpora included around 80 MB of data, and 1,044,764 sentences constructed from 11.67M of words that were not tokenized. The data was introduced as pure text encoded in UTF-8. Furthermore, texts were separated into sentences (one per line) and structured in language pairs. The vocabulary consisted of 148,170 unique Polish and 109,326 unique English word forms. The disproportionate vocabulary size and number of words in each of the corpora presented a challenge, especially when it came to translation from English to Polish .

Before the training of the translation model, pre-processing had to be performed. This included the removal of long sentences (set to 80 tokens) using the Moses toolkit scripts [9]. Moses is an open-source toolkit for statistical machine translation, which supports linguistically motivated factors, confusion network decoding, and efficient data formats for translation models and language models. In addition to the SMT decoder, the toolkit also includes a wide variety of tools for training, tuning and applying the system to many translation tasks.

### 3.1. Polish stem extraction

In order to cope with the vocabulary size discrepancy, as previously mentioned, stems extracted from Polish words were used instead of surface forms. Since the target language was English in the form of normal sentences, it was not necessary to introduce models for converting the stems to the appropriate grammatical forms. For Polish stem extraction, a set of natural language processing tools available at http://nlp.pwr.wroc.pl was used [8]. These tools can be used for:

- Tokenization

- Morphosyntactic analysis

- Text transformation into the featured vectors

The following two components were also included:

- MACA-a universal framework used to connect the different morphological data

- WCRFT - this framework combines conditional random fields and tiered tagging

These tools used sequentially provide output in XML format. This includes the surface form of the to- kens, stems and morphosyntactic tags. With these les we were able to convert polish words to their surface forms and change the word order in sentences to meet the SVO order. This step was relevant because in English the sentence is usually structured according to the Subject-Verb-Object (SVO) word order. Contrariwise, in Polish the word order is much more relaxed and in itself has no significant impact on the sentence meaning, which is decided mostly around the complicated inflection instead.

### 3.2. English data preparation

The English data preparation was far less complex than that of the Polish data. We applied a tool to clean the English data by eliminating foreign words, strange symbols, etc. Compared to Polish, the English data included significantly fewer errors. However, some problems needed to be fixed. The most problematic were portions of text in other languages and strange Unicode symbols.

## 4. ASR techniques

The goal of ASR is to convert a sequence of acoustic observations into a sequence of words. This is typically achieved by a Markov process which uses a set of observable acoustic probabilities to derive a sequence of hidden states which form a hierarchal structure that models everything from phonetics to word context in the language.

The system used in this paper is based on the Kaldi Speech Recognition toolkit [10], which is a WFST based ASR engine containing many of the state-of- the-art methods and algorithms for ASR, including: cepstral-mean normalization (CMN), linear discriminant analysis (LDA) and maximum likelihood linear transformation (MLLT) feature transformation, vocal tract length normalization (VTLN), subspace Gaussian mixture modelling (SGMM), feature space maximum likelihood linear regression (fMLLR) adaptation, training using maximum mutual information (MMI) criterion and speaker adaptive training (SAT), artificial neural network (ANN) and deep neural net- work (DNN) based models. The system is capable of both offline and online decoding and can output results in various formats including 1-best, n-best and lattices.

| Method | WER |
|---|---|
| Inital trigram | 37.37 |
| +LDA/MLLT | 34.37 |
| +MMI | 32.01 |
| +MPE | 32.55 |
| +SAT(fMLLR) | 33.51 |
| +MMI | 31.81 |
| +fMMI | 29.85 |
| +BMMI | 29.69 |
| +SGMM | 32.39 |

Table I. Comparison of training methods in Kaldi on an sample parliamentary training set.

All the experiments were performed on corpora sampled at 16kHz, saved as uncompressed 16-bit little-endian WAV audio les. The acoustic model was trained with 13 MFCC features (including C0) with delta and acceleration coefficients, giving 39 input features per frame. The signal was framed at 100 frames per second with each frame being 25ms long.

The standard Kaldi training procedure consists of several stages, each algorithm building on top of results obtained in the previous stage. For an example training set on the Polish Parliament data, the results shown in table I were achieved.

The initial model consists of a standard left-to- right 5-state triphone model. It is also normalized using Cepstral Mean and Variance Normalization. In the next experiment, an LDA transform is calculated on the input data. Following that, two types of optimization mechanisms are tested, with Maximum- Mutual Information criterion performing somewhat better than Minimum Phone Error. Next, speaker adaptation is performed using feature space Maxi- mum Likelihood Linear Regression, after which the MMI adaptation performs even better. Several types of MMI methods are tested: the standard MMI is improved by optimizing both model and feature space errors in the fMMI technique, while BMMI additionally boosts paths that contain more errors. The SGMM method is an optimization technique where all the phonetic states share a common GMM structure. This can prove beneficial in situations when small amounts of training data is available, but in our case it is al- ways outperformed by the aforementioned methods.

Several iterations of systems were tested, with most improvements achieved by adding data for language modeling and increasing vocabulary size. In fact, vocabulary size was the main cause of errors in the experiments, with Out-Of-Vocabulary (OOV) words causing most issues. Results for various domains are presented in table II.

## 5. SMT techniques

A number of experiments were performed to evaluate various versions for our SMT systems. The experiments involved a number of steps. Processing of the corpora was performed, which included: tokenization, cleaning, factorization, conversion to lower case, splitting, and a final cleaning after splitting. Training data was then processed in order to develop the language model. Tuning was performed for each experiment. Lastly, the experiments were conducted and evaluated using a series of metrics.

| Domain | WER | Vocabulary size |
|---|---|---|
| Polish Senate and Parilament | 19.6 | ~86.5k |
| Television news | 15.72 | ~42k |
| Lectures | 27.75 | ~210k |

Table II. Comparison speech recognition performance for different domains.

The baseline system testing was done using the Moses open source SMT toolkit with its Experiment Management System (EMS) [11]. The SRI Language Modeling Toolkit (SRILM) [12] with an interpolated version of the Kneser-Key discounting (interpolate & kndiscount) was used for 5-gram language model training. We used the MGIZA++ tool for word and phrase alignment. KenLM [13] was used to binarize the language model, with a lexical reordering set to use the msd-bidirectional-fe model. Reordering probabilities of phrases was conditioned on lexical values of a phrase. This considers three different orientation types on source and target phrases like mono- tone(M), swap(S) and discontinuous(D). The bidirectional reordering model adds probabilities of possible mutual positions of source counterparts to current and following phrases. The probability distribution to a foreign phrase is determined by f and to the English phrase by e . MGIZA++ is a multi-threaded version of the well-known GIZA++ tool. The symmetrisation method was set to grow-diag-final-and for word alignment processing. First two-way direction alignments obtained from GIZA++ were intersected, so only the alignment points that occurred in both alignments remained. In the second phase, additional alignment points existing in their union were added. The growing step adds potential alignment points of unaligned words and neighbours. Neighbourhood can be set directly to left, right, top or bottom, as well as to diagonal (grow-diag). In the final step, alignment points between words from which at least one is unaligned were added (grow-diag-final). If the grow-diag-final-and method is used, an alignment point between two unaligned words appears.

## 5.1. TED results

The experiments on TED, conducted with the use of the test data from years 2010-2013, are de ned in table III and Table IV, respectively, for the Polish to English and English to Polish translations. They are measured by the BLEU, NIST, TER and METEOR metrics [14]. Note that a lower value of the TER metric is better, while the other metrics are better when their values are higher. BASE stands for baseline sys- tem with no improvements, COR is a system with corrected spelling in Polish data, INF is a system using infinitive forms in Polish, SVO is a system with the subject-verb-object word order in a sentence and BEST stands for the best result we achieved.

| System | Year | BLEU | NIST | TER |
| --- | --- | --- | --- | --- |
| BASE | 2010 | 16.02 | 5.28 | 66.49 |
| COR | 2010 | 16.09 | 5.22 | 67.32 |
| BEST | 2010 | 20.88 | 5.70 | 64.39 |
| INF | 2010 | 13.22 | 4.74 | 70.26 |
| SVO | 2010 | 9.29 | 4.37 | 76.59 |
| BASE | 2011 | 18.86 | 5.75 | 62.70 |
| COR | 2011 | 19.18 | 5.72 | 63.14 |
| BEST | 2011 | 23.70 | 6.20 | 59.36 |
| BASE | 2012 | 15.83 | 5.26 | 66.48 |
| POPR | 2012 | 15.86 | 5.32 | 66.22 |
| BEST | 2012 | 20.24 | 5.76 | 63.79 |
| BASE | 2013 | 16.55 | 5.37 | 65.54 |
| COR | 2013 | 16.98 | 5.44 | 65.40 |
| BEST | 2013 | 23.00 | 6.07 | 61.12 |
| INF | 2013 | 12.40 | 4.75 | 70.38 |

Table III. Polish-to-English translation of TED.

| System | Year | BLEU | NIST | TER |
| --- | --- | --- | --- | --- |
| BASE | 2010 | 8.49 | 3.70 | 76.39 |
| POPR | 2010 | 9.39 | 3.96 | 74.31 |
| BEST | 2010 | 10.72 | 4.18 | 72.93 |
| INF | 2010 | 9.11 | 4.46 | 74.28 |
| SVO | 2010 | 4.27 | 4.27 | 76.75 |
| BASE | 2011 | 10.77 | 4.14 | 71.72 |
| POPR | 2011 | 10.74 | 4.14 | 71.70 |
| BEST | 2011 | 15.62 | 4.81 | 67.16 |
| BASE | 2012 | 8.71 | 3.70 | 78.46 |
| POPR | 2012 | 8.72 | 3.70 | 78.57 |
| BEST | 2012 | 13.52 | 4.35 | 73.36 |
| BASE | 2013 | 9.35 | 3.69 | 78.13 |
| POPR | 2013 | 9.35 | 3.70 | 78.10 |
| BEST | 2013 | 14.37 | 4.42 | 72.06 |
| INF | 2013 | 13.30 | 4.83 | 70.50 |

Table IV. English-to-Polish translation of TED.

## 5.2. EuroParl and OpenSubtitles results

For the EuroParl and the OPUS OpenSubtitles we conducted experiments on phrase-based system as well as factored system enriched with POS tags. The use of compound splitting and true casing was optional. Some language models were chosen based on their perplexity measure and then also linearly interpolated. Table V shows partial results of our experiments. We used shortcuts abbreviations E (EuroParl) and O (OpenSubtitles), if there is no additional suffix it means that test was baseline system trained on phrase-based model, suffix F (e.g. TF) means we used factored model, T refers to data that was true-cased and C means that a compound splitter was used. If the suffix is I we used infinitive forms of all polish data and the S suffix refers to changes in word order to meet SVO schema. In EuroParl experiments suffix L stands for bigger EN in-domain language model. H stands for highest score we obtained by combining methods and interpolating extra data. G su x stands for tests on translation of our data by Google Translator.

|     | EN->PL BLEU | EN->PL NIST | EN->PL TER | PL->EN BLEU | PL->EN NIST | PL->EN TER |
| --- | --- | --- | --- | --- | --- | --- |
| E   | 73.18 | 11.79 | 22.03 | 67.71 | 11.07 | 25.69 |
| EL  | 80.60 | 12.44 | 12.44 | -     | -     | -     |
| ELC | 80.68 | 12.46 | 16.78 | 67.69 | 11.06 | 25.68 |
| ELT | 78.09 | 12.41 | 17.09 | 64.50 | 10.99 | 26.28 |
| ELF | 80.42 | 12.44 | 17.24 | 69.02 | 11.15 | 24.79 |
| ELI | 70.45 | 11.49 | 23.54 | 70.73 | 11.44 | 22.50 |
| ELS | 61.51 | 10.65 | 31.71 | 49.69 | 9.38  | 40.51 |
| ELH | 82.48 | 12.63 | 15.73 | -     | -     | -     |
| EG  | 32.87 | 7.57  | 50.57 | 22.95 | 6.01  | 46.33 |
| O   | 53.21 | 7.57  | 46.01 | 51.87 | 7.04  | 47.66 |
| OC  | 53.13 | 7.58  | 45.70 | -     | -     | -     |
| OT  | 52.63 | 7.58  | 45.01 | 50.57 | 6.91  | 48.43 |
| OF  | 53.51 | 7.61  | 45.70 | 52.01 | 6.97  | 48.22 |
| OG  | 22.98 | 4.76  | 68.21 | 16.36 | 3.69  | 77.01 |

Table V. Results EuroParl and OpenSubtitles.

### 5.3. EMEA results

For the EMEA corpora we conducted 12 experiments as shown in table VI and table VII. The experiment 00 in these tables illustrates the baseline system. Every single experiment comes as a separate modification of the baseline. In addition, the Experiment 01 relies on the true casing and punctuation normalization.

The Experiment 02 is improved with the help of the Operation Sequence Model (OSM). The reason for the introduction of the OSM is the provision of phrase-based SMT models, which can memorize dependencies and lexical triggers. In addition, the OSM uses a source and target context, with an exclusion of the spurious phrasal segmentation problems. The OSM is invaluable, especially for the strong mechanisms of reordering. It combines both translation and reordering, deals with the short and long-distance re- ordering, and does not ask for a reordering limit [15].

The Experiment 03 includes a factored model that ensures an additional annotation on the word levels, with a possibility to be exploited in different models. We evaluate the part of speech tagged data in correlation with the English language segment as a basis for the factored phrase models [16].

Hierarchically structured phrase-based translations include both the strengths of phrase-based and syntax-based translations. They use phrases (word segments or blocks) as translation units, including the synchronous context-free grammar cases as rules (syntax-based translations). Hierarchically structured phrase models ensure the rules with gaps. Since these are illustrated by non-terminals and the rules them- selves are best evaluated with a search algorithm, which is similar to syntactic chart parsing. These models can be categorized as the class of tree-based or grammar-based models. We applied such a model in the Experiment 04.

The Target Syntax model includes the application of linguistic annotation for non-terminals in the hierarchically structured models. This asks for a syntactic parser. In this case, we applied the Collins [17] statistical parser of the natural language in the Experiment 05.

The Experiment 06 was executed with the help of stemmed alignment of words. The factored translation model enabled it to come up with the word alignment based on word structure, which differs from the surface word formations. One, apparently very popular, method is to apply stemmed words for these alignments of words. There are two main reasons for such actions. For morphologically richness of languages, stemming deals with the data parity problem. On the other hand, GIZA++ may face serious challenges with the immense vocabulary, while the stemming influences the number of unique words.

The Experiment 07 applies Dyer's Fast Align [18], which is actually an alternative for the GIZA++. It works much faster with the better results, especially for language pairs, which come with no need for the large-scale reordering.

In the Experiment 08 we applied settings recommended by Koehn with his system of Statistical Ma- chine Translation in WMT'13 [19].

In Experiment 09 we changed the language model discounting to Witten-Bell. This discounting method considers diversity of predicted words. It was developed for text compression and can be considered to be an instance of Jelinek-Mercer smoothing. The n-th order smoothed model is de ned recursively as a linear interpolation between the n-th order maximum likelihood model and the (n-1)th order smooth model [20].

Lexical reordering was set to hier-mslr-bidirectional-fe in Experiment 10. It is a hierarchical reordering method that considers different orientations: monotone, swap, discontinuous-left, and discontinuous-right. The reordering is modelled bidirectionally, based on the previous or next phrase, conditioned on both the source and target languages.

Compounding is a method of word formation consisting of a combination of two (or more) autonomous lexical elements that form a unit of meaning. This phenomenon is common in German, Dutch, Greek, Swedish, Danish, Finnish, and many other languages.

| System | BLEU | NIST | TER |
|---|---|---|---|
| 00 | 70.15 | 10.53 | 29.38 |
| 01 | 64.58 | 9.77 | 35.62 |
| 02 | 71.04 | 10.61 | 28.33 |
| 03 | 71.22 | 10.58 | 28.51 |
| 04 | 76.34 | 10.99 | 24.77 |
| 05 | 70.33 | 10.55 | 29.27 |
| 06 | 71.43 | 10.60 | 28.73 |
| 07 | 71.91 | 10.76 | 26.60 |
| 08 | 71.12 | 10.37 | 29.95 |
| 09 | 71.32 | 10.70 | 27.68 |
| 10 | 71.35 | 10.40 | 29.74 |
| 11 | 70.34 | 10.64 | 28.22 |
| 12 | 72.51 | 10.70 | 28.19 |

Table VI. Polish-to-English results of EMEA.

| System | BLEU | NIST | TER |
|---|---|---|---|
| 00 | 69.18 | 10.14 | 30.39 |
| 01 | 61.15 | 9.19 | 39.45 |
| 02 | 69.41 | 10.14 | 30.90 |
| 03 | 68.45 | 10.06 | 31.62 |
| 04 | 73.32 | 10.48 | 27.05 |
| 05 | 69.21 | 10.15 | 30.88 |
| 06 | 69.27 | 10.16 | 31.27 |
| 07 | 68.43 | 10.07 | 33.05 |
| 08 | 67.61 | 9.87 | 29.95 |
| 09 | 68.98 | 10.11 | 31.13 |
| 10 | 68.67 | 10.02 | 31.92 |
| 11 | 69.01 | 10.14 | 30.84 |
| 12 | 67.47 | 9.89 | 33.32 |

Table VII. English-to-Polish results of EMEA.

For example, the word "flowerpot" is a closed or open compound in English texts. This results in a lot of unknown words in any text, so splitting up these com- pounds is a common method when translating from such languages. Moses offers a support tool that splits up words if the geometric average of the frequency of its parts is higher than the frequency of a word. In Experiment 11 we used the compound splitting feature. Lastly, for Experiment 12 we used the same settings as for out of domain corpora in IWSLT'13 [11].

**5.4. Speech-to-speech results**

Previous experiments dealt with ASR and SMT performance as individual systems, but ultimately the system was to be used in a speech-to-speech manner, meaning we would get audio as input and expect text and/or audio in a different language as output. The errors of that process wouldn't be merely a sum of errors of underlying tasks, because the errors in previous steps would cause unforeseen consequences to the following. For example, a word missing in ASR output would cause the SMT system to under-perform in un- predictable ways. Another issue is caused by the fact that SMT expects fully punctuated and capitalized text, but ASR returns only a string of words in lower case.

| Data set   | BLEU  |
|------------|-------|
| Original   | 36.52 |
| Normalized | 25.12 |
| ASR output | 22.94 |

Table VIII. Comparison of translation results for various outputs of the S2S system.

To measure the performance of the SMT system we prepared three sets of 24 sentences chosen from the development set of the Euronews corpus: the original hand-proofed transcript, the normalized, non punctuated and lowercase version of the same transcript and finally the result of ASR. This was com- pared to the hand proofed translation of these sentences. The performance of the ASR on these sentences was 10.7% WER.

## 6. System setup

Both the SMT and ASR systems discussed in previous chapters have the ability to perform online computation and work in near real-time manner. A couple of demonstration applications were therefore prepared to present the system in real-life situations.

The back-end system was designed in two different ways, according to the requirements of the U-Star and EU-Bridge projects accordingly.

The U-Star project developed s set of standardized protocols specifically for S2S translation. The proto- col is based around an XML-like markup language named MCML (Modality Conversion Markup Lan- guage) whose purpose is to convert information from one modality (usually speech or text in a particular language) to another. Requests in the system function as REST calls with MCML documents serving as the basis of communication. The system architecture is based on a series of Tomcat servers that route the in- formation requests from the clients to the respective engines. The organization of the servers is tree-like in structure and allows to perform pivoting , ie. if a language pair is not directly translatable by any of the available engines, a third language can be utilized to complete the transaction.

The EU-Bridge project adapted a protocol from one of its early members, that was already used in a commercial program for mobile S2S translation. It uses a much more lightweight XML-like serialized protocol than the complicated MCML document. The overall structure of the server is also much more centralized, ie. star-shaped. There is a single point of convergence for all the participants (clients and engines) all of which share a similar (but slightly different) proto- col to the central server.

To test the two service architectures, an application was created for the Android operating system. The application features a text input and a text output field, with the ability to switch languages on either of the fields. The input also has the ability to accept recorded audio.

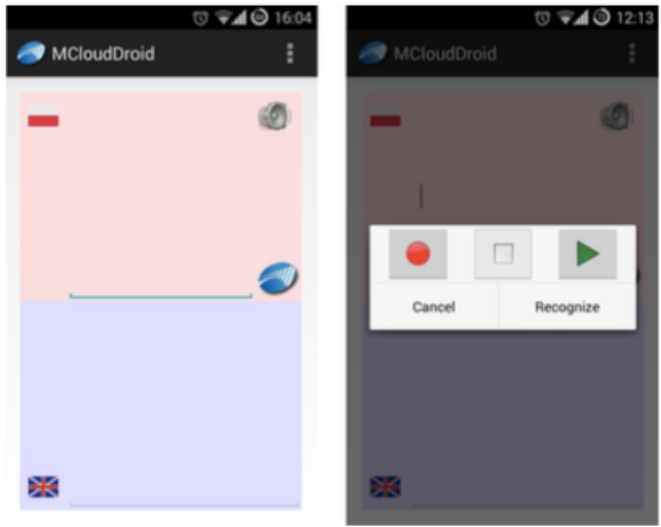

Figure 1. Screenshots of the sample Android application. Shows the main screen of the application (left) and the voice input dialog (right).

## 7. Results and conclusions

After the transcripts have been extracted from the supplied XML les, the same number of lines for both languages was obtained, but there were some discrepancies in the parallel text. Those differences were caused mostly by repetitions in the Polish text and some additional remarks (like Applause or Thanks ) which were not present in the English text. 28 lines have been manually corrected for the whole set of 134325 lines. Given that the talks were translated mostly by volunteers, the training material seemed a bit noisy and imprecise at times.

The vocabulary sizes (extracted using SRILM) were 198622 for Polish and 91479 for English, which shows the fundamental problem for the translation - a huge difference in the vocabulary size.

Tokenization of the input data was done using standard tools delivered with Moses with an extension done by FBK for Polish.

Before training a translation model, the usual pre- processing was applied, such as removing long sentences (threshold 60) and sentences with length difference exceeding a certain threshold. This was again done using scripts from the Moses toolkit.

The final tokenized, lowercased and cleaned training corpora for Polish and English contained 132307 lines, but with an even bigger difference in thr vocabulary size - 47250 for English vs. 123853 for Polish.

This big discrepancy between source and target vocabularies shows a necessity of using additional knowledge sources. First, we decided to limit the size of the Polish vocabulary by using stems instead of surface forms and, second, to use morphosyntactic tagging as an additional source of information for the preparation of the SMT system.

Several conclusions can be drawn from the experiment results presented here. It was surprising that the truecasing and punctuation normalization decreased the scores by a significant factor. We suppose that the text was already properly cased and punctuated. We observed that, quite strangely, OSM decreased some metrics' results. It usually increases the translation quality. However, in the PL->EN experiment the BLEU score increased just slightly, but other metrics decreased at the same time. A similar phenomenon can be seen in the EN->PL experiments. Here, the BLEU score increased, but other metrics decreased.

Most of the other experiments worked as anticipated. Almost all of them raised the score a little bit, or were at least confirmed, with each metric increasing in the same manner. Unfortunately, replication of the settings that provided the best system score on IWSLT 2013 evaluation campaign, did not improve the quality on each data set as we had hoped. The most likely reason is that the data used in the IWSLT did not come from any specific text domain, while elsewhere we dealt with very narrow domains. It may also mean that training and tuning parameter adjustment may be required separately for each text domain if improvements cannot be simply replicated.

On the other hand, improvements obtained by training the hierarchical based model were surprising. The same, significant improvements can be observed in both the PL->EN and EN->PL translations, which most likely provide a very good starting point for future experiments.

Translation from EN to PL is more di cult, which is demonstrated by the worse results obtained in the experiments. The most likely cause is the complicated Polish grammar as well as the larger vocabulary size.

The analysis of our experiments led us to conclude that the results of the translations, in which the BLEU measure is greater than 70, can be considered very good within the selected text domain and those above 50 satisfactory. The high evaluation scores indicate that the translations of some of the tested systems should be understandable by a human and good enough to help him in his work. We strongly believe that improving the BLEU score to a threshold over 80 or even 85 would produce systems that could be used in practical applications, when it comes to PL- EN translations. It may be particularly helpful with the Polish language, which is complex from many aspects of its structure, grammar and spelling.

Finally, the preliminary S2S experiments presented a few outstanding issues to overcome. The greatest of them being the lack of sentence boundaries, punctuation and capitalization. ASR generally outputs only a stream of words and the lack of any sentence boundaried breaks the SMT process. The actual word errors introduced by the ASR do present a problem, but, de- pending on the domain, they needn't be too harmful and the system can still be quite functional. If the error rate is fairly small, the translation is pretty legible, with OOV being the key issue - a rare word or an entity name missing from the ASR output would be unrecoverable in the latter stages of the process. Disregarding the problems for a moment, the system works surprisingly well at times. The authors hope to evolve it into a fully functional system capable of serving as an agent in a conversation between two humans speaking in different languages.

**Acknowledgement**

The research leading to these results has received funding from the European Union Seventh Framework Programme (FP7/2007-2013) under grant agreement no 287658.

**References**

[1] Marcin Miakowski, Georg Rehm, and Hans Uszkoreit, The Polish language in the digital age, Springer, 2012.

[2] Krzysztof Marasek, Łukasz Brocki, Danijel Korzinek, Krzysztof Szklanny, and Ryszard Gubrynowicz, User-centered design for a voice portal, in Aspects of Natural Language Processing, pp. 273 293. Springer, 2009.

[3] Grażyna Demenko, Stefan Grocholewski, Katarzyna Klessa, Jerzy Ogórkiewicz, Agnieszka Wagner, Marek Lange, Daniel Sledzinski, and Natalia Cylwik, Jurisdic: Polish speech database for taking dictation of legal texts., in LREC, 2008.

[4] Krzysztof Marasek, Ted polish-to-english translation system for the iwslt 2012, Proceedings IWSLT 2012, 2012.

[5] Xuedong Huang, James Baker, and Raj Reddy, A historical perspective of speech recognition, Communications of the ACM, vol. 57, no. 1, pp. 94 103, 2014.

[6] Dorota J Iskra, Beate Grosskopf, Krzysztof Marasek, Henk van den Heuvel, Frank Diehl, and Andreas Kiessling, Speecon-speech databases for consumer devices: Database specification and validation., in LREC, 2002.

[7] Adam Przepiórkowski, Korpus ipi pan, Wersja wstepna. Instytut Podstaw Informatyki, Polska Akademia Nauk, Warszawa, 2004.

[8] Krzysztof Wołk and Krzysztof Marasek, Real-time statistical speech translation, in New Perspectives in Information Systems and Technologies, Volume 1, Álvaro Rocha, Ana Maria Correia, Felix . B Tan, and Karl . A Stroetmann, Eds., vol. 275 of Advances in Intelligent Systems and Computing, pp. 107 113. Springer International Publishing, 2014.

[9] Philipp Koehn, Hieu Hoang, Alexandra Birch, Chris Callison-Burch, Marcello Federico, Nicola Bertoldi, Brooke Cowan, Wade Shen, Christine Moran, Richard Zens, et al., Moses: Open

source toolkit for statistical machine translation, in Proceedings of the 45th Annual Meeting of the ACL on Interactive Poster and Demonstration Sessions. Association for Computational Linguistics, 2007, pp. 177 180.

[10] Daniel Povey, Arnab Ghoshal, Gilles Boulianne, Lukas Burget, Ondrej Glembek, Nagendra Goel, Mirko Hannemann, Petr Motlicek, Yanmin Qian, Petr Schwarz, et al., The kaldi speech recognition toolkit, in Proc. ASRU, 2011, pp. 1 4.

[11] Krzysztof Wołk and Krzysztof Marasek, Polish â english speech statistical machine translation systems for the iwslt 2013, in Proceedings of the 10th International Workshop on Spoken Language Translation, 2013, pp. 113 119.

[12] Andreas Stolcke et al. „Srilm-an extensible language modeling toolkit", in INTERSPEECH, 2002.

[13] Kenneth Heafield, Clark, and Philipp Koehn, Scalable modified Kneser- Ney language model estimation, in Proceedings of the 51st Annual Meeting of the Association for Computational Linguistics, So a, Bulgaria, August 2013, pp. 690 696.

[14] Krzysztof Wołk and Krzysztof Marasek, Alignment of the polish-english parallel text for a statistical machine translation, Computer Technology and Application, pp. 575 583, 2013.

[15] Nadir Durrani, Alexander Fraser, Helmut Schmid, Hassan Sajjad, and Richárd Farkas, Munich- edinburgh-stuttgart submissions of osm systems at wmt13, in Proceedings of the Eighth Workshop on Statistical Machine Translation, 2013, pp. 120 125.

[16] Philipp Koehn and Hieu Hoang, Factored translation models., in EMNLP-CoNLL. Citeseer, 2007, pp. 868 876.

[17] Daniel M Bikel, Intricacies of collins' parsing model, Computational Linguistics, vol. 30, no. 4, pp. 479 511, 2004.

[18] Chris Dyer, Victor Chahuneau, and Noah A Smith, A simple, fast, and effective reparameterization of ibm model 2., in HLT-NAACL. Citeseer, 2013, pp. 644 648.

[19] Ondrej Bojar, Christian Buck, Chris Callison-Burch, Christian Federmann, Barry Haddow, Philipp Koehn, Christof Monz, Matt Post, Radu Soricut, and Lucia Specia, Findings of the 2013 workshop on statistical machine translation, in Proceedings of the Eighth Workshop on Statistical Machine Translation, 2013, pp. 1 44.

[20] AS M Mahmudul Hasan, Saria Islam, and M Arifur Rahman, A comparative study of witten bell and kneser-ney smoothing methods for statistical machine translation, Journal of Information Technology, vol. 1, pp. 1 6, 2012.